\documentclass[fleqn,10pt]{wlscirep}
\usepackage[utf8]{inputenc}
\usepackage[T1]{fontenc}
\usepackage[numbers,comma,sort&compress]{natbib} 
\usepackage{lineno}
\usepackage{xspace}
\usepackage[most]{tcolorbox} 
\tcbuselibrary{listingsutf8} 
\usepackage{mdframed}
\usepackage{enumitem}

\usepackage{graphicx}%
\usepackage{xurl}%
\usepackage{multirow}%
\usepackage{amsmath,amssymb,amsfonts}%
\usepackage{amsthm}%
\usepackage{mathrsfs}%
\usepackage[title]{appendix}%
\usepackage{xcolor}%
\usepackage{textcomp}%
\usepackage{manyfoot}%
\usepackage{booktabs}%
\usepackage{xurl}%
\usepackage{algorithm}%
\usepackage{algorithmicx}%
\usepackage{algpseudocode}%
\usepackage{listings}%

\usepackage{listings}
\usepackage{float}

\tcbset{
  aibox/.style={
    width=500pt,
    top=10pt,
    colback=white,
    colframe=black,
    colbacktitle=black,
    enhanced,
    center,
    attach boxed title to top left={yshift=-0.1in,xshift=0.15in},
    boxed title style={boxrule=0pt,colframe=white,},
  }
}

\makeatletter
\newtcolorbox[auto counter]{AIbox}[2][]{%
  aibox,
  title={Prompt~\thetcbcounter: #2},
  before upper={\protected@edef\@currentlabel{\thetcbcounter}},
  #1
}
\makeatother

\title{Med-V1: Small Language Models for Zero-shot and Scalable Biomedical Evidence Attribution}

\author[1]{Qiao Jin}
\author[1]{Yin Fang}
\author[1]{Lauren He}
\author[1]{Yifan Yang}
\author[2]{Guangzhi Xiong}
\author[1]{Zhizheng Wang}
\author[1]{Nicholas Wan}
\author[1]{Joey Chan}
\author[1]{Donald C. Comeau}
\author[1]{Robert Leaman}
\author[3]{Charalampos S. Floudas}
\author[2]{Aidong Zhang}
\author[1,4]{Michael F. Chiang}
\author[5,6]{Yifan Peng}
\author[1,+]{Zhiyong Lu}

\affil[1]{Division of Intramural Research, National Library of Medicine, National Institutes of Health}
\affil[2]{Department of Computer Science, University of Virginia}
\affil[3]{Center for Cancer Research, National Cancer Institute, National Institutes of Health}
\affil[4]{National Eye Institute, National Institutes of Health}
\affil[5]{Department of Population Health Sciences, Weill Cornell Medicine}
\affil[6]{Institute of AI for Digital Health, Weill Cornell Medicine}

\affil[+]{Corresponding author (\url{zhiyong.lu@nih.gov}).}

\begin{abstract}
Assessing whether an article supports an assertion is essential for hallucination detection and claim verification. While large language models (LLMs) have the potential to automate this task, achieving strong performance requires frontier models such as GPT-5 that are prohibitively expensive to deploy at scale. To efficiently perform biomedical evidence attribution, we present Med-V1, a family of small language models with only three billion parameters. Trained on high-quality synthetic data newly developed in this study, Med-V1 substantially outperforms (+27.0\% to +71.3\%) its base models on five biomedical benchmarks unified into a verification format. Despite its smaller size, Med-V1 performs comparably to frontier LLMs such as GPT-5, along with high-quality explanations for its predictions. We use Med-V1 to conduct a first-of-its-kind use case study that quantifies hallucinations in LLM-generated answers under different citation instructions. Results show that the format instruction strongly affects citation validity and hallucination, with GPT-5 generating more claims but exhibiting hallucination rates similar to GPT-4o. Additionally, we present a second use case showing that Med-V1 can automatically identify high-stakes evidence misattributions in clinical practice guidelines, revealing potentially negative public health impacts that are otherwise challenging to identify at scale. Overall, Med-V1 provides an efficient and accurate lightweight alternative to frontier LLMs for practical, real-world biomedical evidence attribution. Med-V1 is available at \url{https://github.com/NLM-DIR/Med-V1}.
\end{abstract}

\begin{document}

\flushbottom
\maketitle

\section{Introduction}\label{introduction}
Biomedical evidence attribution is the task of determining whether a biomedical source supports an assertion, contradicts it, or provides insufficient evidence to assess it \cite{vladika-matthes-2023-scientific}.
A common example is automatic fact-checking. Such systems usually retrieve evidence relevant to a given claim and then apply a verification model to determine the claim's veracity based on that evidence \cite{guo2022survey, wadden-etal-2022-scifact, petroni2023improving}.
The need for robust verification has grown with the increasing use of generative artificial intelligence (AI) tools such as ChatGPT, which are known to hallucinate when generating statements with citations \cite{zuccon2023chatgpt, liu-etal-2023-evaluating, wu2025automated}.
As a result, their outputs must be thoroughly verified before downstream usage \cite{jin2023retrieve, augenstein2024factuality, asai2024openscholar}, especially in high-stakes domains such as healthcare \cite{wang-etal-2025-medcite}.
While frontier large language models (LLMs) have shown strong performance across various biomedical tasks \cite{thirunavukarasu2023large, clusmann2023future, tian2023opportunities, wang2023pre, shah2023creation, he2024foundation, omiye2024large, liu2025application} such as question answering \cite{singhal2023large, nori2023can, chen2024effect} and clinical trial matching \cite{wong2023scaling, jin2024matching, wornow2025zero}, they often rely on commercial systems and are prohibitively expensive for routine, large-scale deployment.
With the rapid growth of AI-generated statements, it is conceivable that verification becomes a heavily used utility task. In practice, such large-scale verification can only be realistically performed by small language models \cite{wang2025survey}, which are more cost-effective and scalable.
We use biomedical evidence attribution as the umbrella term for this assertion-source evaluation. Within this framework, claim verification refers to evaluating an independently stated factual claim, whereas citation attribution refers to evaluating a statement against its cited source. Binary question answering can be formulated in the same framework by converting a yes/no question and answer into a declarative assertion.
As such, the objective of this work is to develop a lightweight foundation model that can perform biomedical evidence attribution comparably with frontier LLMs, generalize well to unseen data distributions, and provide faithful explanations for its predictions.

To this end, we first introduce a novel LLM-driven pipeline that generates MedFact-Synth, a large-scale synthetic dataset comprising biomedical claims and their high-quality verification reports.
Using MedFact-Synth, we subsequently fine-tune small LLMs with only three billion parameters \cite{grattafiori2024llama3herdmodels, qwen2025qwen25technicalreport}, resulting in Med-V1.
For evaluation, we compile MedFact-Bench, which unifies five biomedical benchmarks spanning claim verification, citation attribution, and repurposed binary question answering in a common assertion-source evaluation format \cite{wadden2020fact, gupta2025dataset, sarrouti2021evidence, jin2019pubmedqa, krithara2023bioasq}.
Experimental results show that Med-V1 substantially outperforms its base models, achieving improvements of 27.0\% to 71.3\% across the datasets.
Despite its small size, Med-V1 reaches performance levels comparable to frontier LLMs such as GPT-5 \cite{openai2025gpt5}.
Further analysis shows that Med-V1 consistently presents high-quality explanations for its predictions.

To demonstrate the real-world utility of Med-V1, we conduct two large-scale use case studies that analyze evidence attribution in both LLM-generated answers and high-stakes clinical practice guidelines. First, we apply Med-V1 to LLM-generated answers under seven citation-instruction formats by extracting claim-citation pairs and verifying each claim against its cited article, which enables large-scale quantification of citation validity and hallucination rates across formats. This analysis shows that citation format strongly affects citation validity and hallucination, and that GPT-5 generates more claims but exhibits hallucination rates similar to its predecessor GPT-4o. Second, we apply Med-V1 to citation statements in recent clinical practice guidelines and identify a nontrivial subset whose cited sources do not support the corresponding claims; after manually validating 100 flagged cases, we find that a significant proportion of them are genuine misattributions, and many involve treatment effects that might lead to potential risks in public health.

In summary, these results show that Med-V1 achieves frontier-level verification performance while remaining compact and computationally efficient. This makes it a practical and accessible alternative to proprietary LLMs for large-scale biomedical evidence attribution, hallucination detection, and citation auditing. 

\section{Results}\label{results}
\subsection{Overview of Med-V1 Training and Inference}
\begin{figure}[h!]
    \centering
    \includegraphics[width=\linewidth]{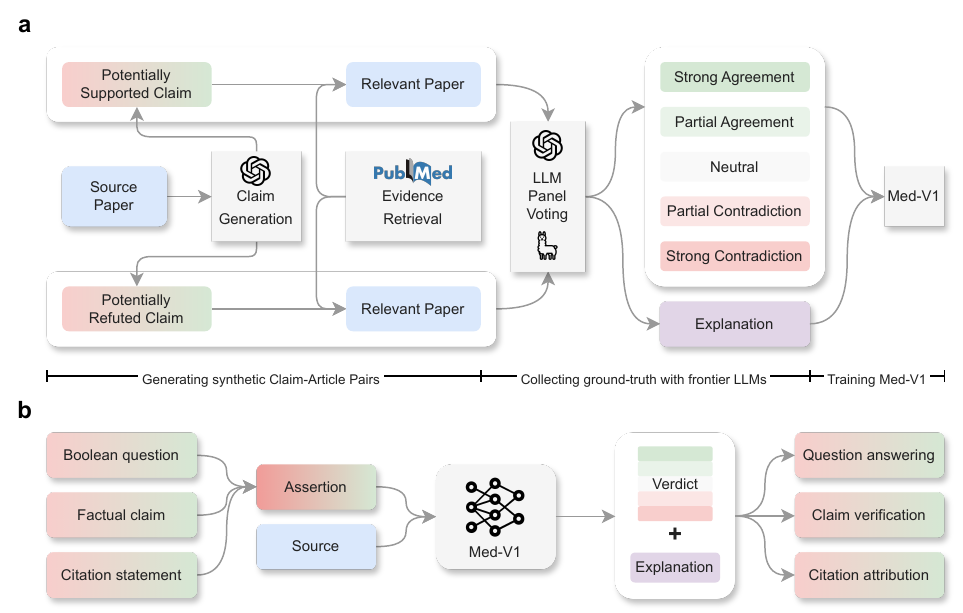}
    \caption{\textbf{Overview of Med-V1 training and inference}. \textbf{a}: MedFact-Synth construction and Med-V1 training. Synthetic claims are generated from source papers and then verified by a panel of LLMs using relevant papers retrieved from PubMed. The resulting verified claim-evidence pairs form the MedFact-Synth dataset, which is then used to train Med-V1 through a combination of supervised fine-tuning and reinforcement learning. \textbf{b}: Inference with Med-V1. Given an assertion and a source biomedical article, Med-V1 assesses whether the article supports the assertion. The assertions can be derived from Boolean questions, factual claims, or citation statements, corresponding to the applications of binary question answering, claim verification, and citation attribution, respectively. We formulate all three applications as biomedical evidence attribution between an assertion and a source article. Med-V1 outputs both a 5-point Likert rating of agreement and a natural-language explanation of its verdict.}
    \label{fig:overview}
\end{figure}

This work contains two phases: (i) synthetic data generation and model training (Figure~\ref{fig:overview}a) and (ii) Med-V1 inference for downstream tasks (Figure~\ref{fig:overview}b).

\textbf{Phase I: Synthetic data generation and model training.} In the first phase, we construct MedFact-Synth, a large-scale synthetic training set including 1.5 million instances. Each instance contains: a synthetic claim to be verified, a source article serving as evidence, a rationale explaining the verification, and a 5-point Likert-scale verdict, ranging from strong contradiction (-2) and partial contradiction (-1) to neutral (0), partial agreement (+1), and strong agreement (+2).
To build this dataset, we begin by sampling one million articles from the PubMed 2025 baseline~\cite{sayers2025database}. 
For each article, GPT-4o-mini \cite{hurst2024gpt} is prompted to generate two claims: one that the article may support and one that it may refute.
To collect diverse claim-article pairs for verification, we use MedCPT~\cite{jin2023medcpt} to retrieve the top 10 most relevant PubMed articles for each claim.
A panel of frontier LLMs then verifies each pair, generating both rationales and verdicts via a voting-based mechanism.
Then, we utilize MedFact-Synth to fine-tune Med-V1, which is a three-billion-parameter (3B) small LLM available in two variants:
Med-V1-L3B, initialized from Llama-3.2-3B-Instruct \cite{grattafiori2024llama3herdmodels} and Med-V1-Q3B, initialized from Qwen2.5-3B-Instruct \cite{qwen2025qwen25technicalreport}.
The quality of MedFact-Synth is examined in the next section, and technical details of the dataset construction and training process are described in the Methods section.

\textbf{Phase II: Inference and downstream applications.} During the inference phase, Med-V1 can be applied to various tasks where an assertion-source pair is available.
Here, an assertion refers to a declarative statement whose factuality needs to be checked, and the source refers to the biomedical article or document against which the assertion is verified.
Assertions may originate from various settings.
For example, boolean questions can be reformulated as declarative assertions, factual claims can be collected on social media, and citation statements can be extracted from AI-generated or human-written content.
When verification sources are not already provided, they can be easily retrieved using an evidence search system \cite{jin2024pubmed, hersh2024search}.
These assertion-source pairs are then fed into Med-V1, which outputs both a structured verdict (agreement level ranging from -2 to +2) and a natural-language explanation (Figure~\ref{fig:overview}b). 
This unified formulation allows Med-V1 to perform multiple downstream applications, including binary question answering, claim verification, and citation attribution.

In this work, we evaluate Med-V1 on standard claim verification datasets and re-purposed binary question answering datasets, under a strict zero-shot setting.
We further validate the quality of Med-V1 explanation, as well as its capability to detect LLM hallucinations and identify citation misattributions in clinical guidelines.

\begin{figure}[ht!]
    \centering
    \includegraphics[width=\linewidth]{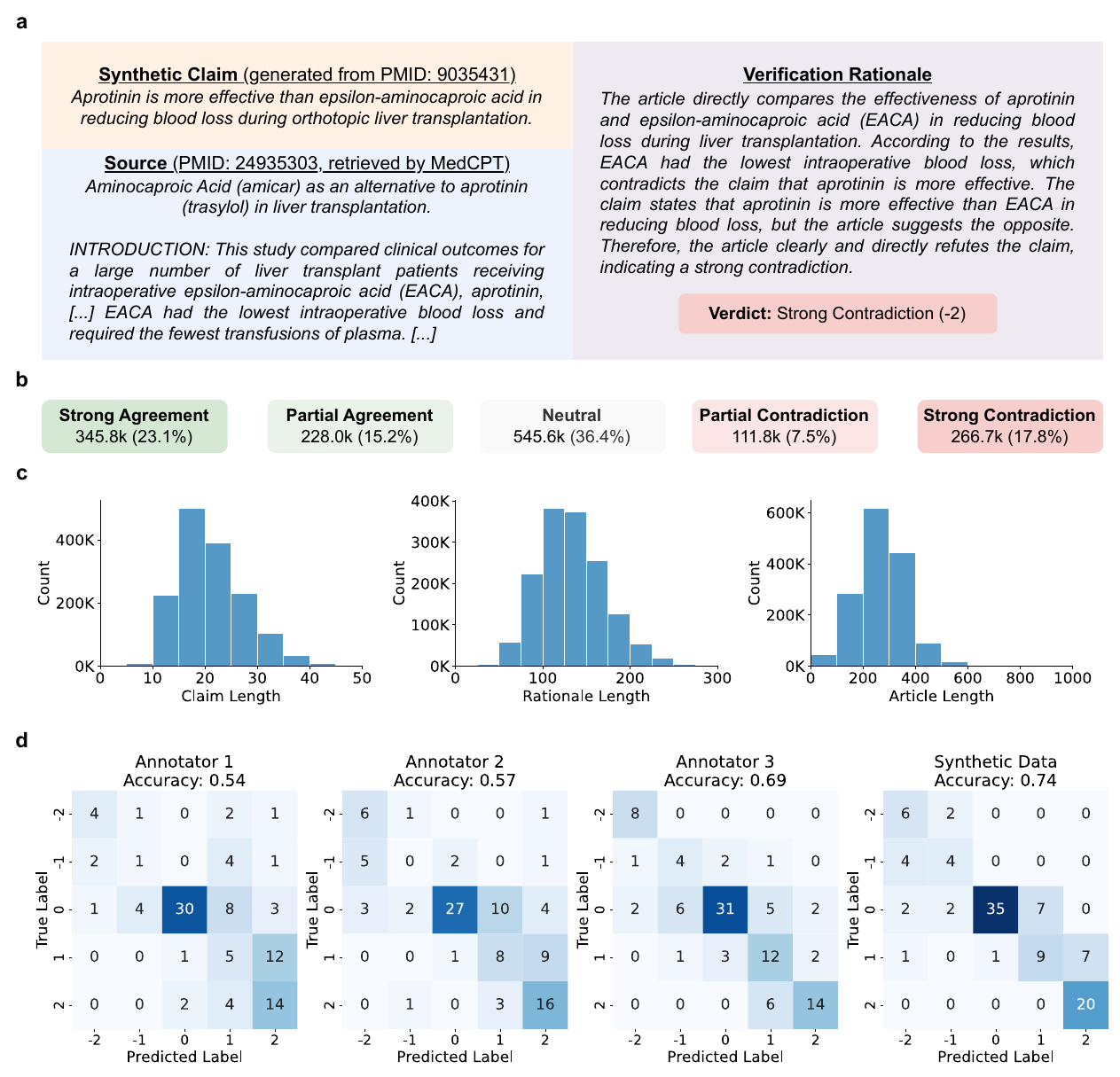}
    \caption{\textbf{Generation and evaluation of MedFact-Synth.} \textbf{a}: An example from MedFact-Synth. \textbf{b}: Distribution of veracity labels in MedFact-Synth. \textbf{c}: Word-count distribution of the claims, rationales, and articles in MedFact-Synth. \textbf{d}: Confusion matrices comparing each annotation method with the true labels, which are the ground-truth derived from annotator consensus. Annotations in MedFact-Synth (Synthetic Data) achieve higher accuracy than those of human annotators.}
    \label{fig:synthetic}
\end{figure}

\subsection{MedFact-Synth is Large-scale and of High-quality}

Figure~\ref{fig:synthetic}a shows an example from the MedFact-Synth dataset. In this example, the claim is generated to be supported by its source article (PMID: 9035431). However, the relevant paper retrieved by MedCPT (PMID: 24935303) actually refutes the claim, as determined by the frontier LLM panel. Because claims are paired with independently retrieved articles rather than being restricted to their source papers, MedFact-Synth naturally includes a mixture of supportive, neutral, and contradictory evidence. As a result, this design leads MedFact-Synth to capture more realistic verification scenarios than settings in which claims are always aligned with their originating article.

Overall, MedFact-Synth contains 1.5 million instances with a reasonably balanced label distribution: 38.3\% (strong and partial) agreement, 36.4\% neutral, and 25.3\% (strong and partial) contradiction (Figure~\ref{fig:synthetic}b). This balance ensures that Med-V1 is exposed to both supportive and conflicting evidence at scale, rather than being dominated by trivially supported claims. The average lengths are 20.7 words for claims, 132.8 words for rationales, and 271.3 words for source articles (Figure~\ref{fig:synthetic}c). Thus, claims remain concise enough to function as standalone assertions, while rationales and articles provide sufficient context for non-trivial reasoning without requiring full-length papers. 

To assess the quality of MedFact-Synth, we randomly sample 100 instances and recruit three annotators (Z.W., N.W., J.C.) with biomedical training to independently evaluate each claim-source pair using the same 5-point Likert scale as MedFact-Synth. Annotators base their judgments solely on the title and abstract of each source article. After the independent annotation round, disagreements are discussed to establish a human consensus label for each instance. The label definitions and annotation procedure are described in the Methods.

Figure~\ref{fig:synthetic}d shows the confusion matrices comparing each annotator and the MedFact-Synth labels with the human consensus labels. The MedFact-Synth label exactly matches the human consensus label in 74 of 100 instances, compared with 54, 57, and 69 instances for the three individual annotators. As a complementary analysis, when the five-point scores are grouped as negative ($-2$/$-1$), neutral ($0$), and positive ($+1$/$+2$), the MedFact-Synth label agrees with the human consensus in 87 of 100 instances, compared with 73, 75, and 78 instances for the three individual annotators.

We further compare the multi-model consensus labels with the three individual LLM labels on the same 100 human-adjudicated instances. Exact five-point agreement with the human consensus is 62\% for GPT-4o-mini, 73\% for Llama-3.3-70B-Instruct, 70\% for o3-mini, and 74\% for the multi-model consensus. The consensus label therefore achieves the highest observed exact agreement in this sample, supporting its use as a higher-quality training target than the labels from any individual model. Because these instances were sampled after consensus filtering, this analysis does not determine how alternative labeling pipelines would affect dataset coverage or downstream benchmark performance (Supplementary Table~4).

Taken together, these analyses show that MedFact-Synth is large-scale and diverse and provide initial evidence that its labels are generally consistent with human judgments. The manual evaluation covers only 100 instances and therefore does not constitute a comprehensive assessment of label quality across the full dataset.

\subsection{Med-V1 Closes the Performance Gap between Frontier and Lightweight LLMs}

To quantitatively measure the performance of Med-V1, we introduce MedFact-Bench. 
It contains five datasets spanning claim verification, citation attribution, and repurposed binary question answering: SciFact \cite{wadden2020fact}, HealthVer \cite{sarrouti2021evidence}, MedAESQA \cite{gupta2025dataset}, PubMedQA-Fact (derived from PubMedQA \cite{jin2019pubmedqa}), and BioASQ-Fact (derived from BioASQ \cite{krithara2023bioasq}). 
Together, MedFact-Bench contains 14,274 claim-article pairs annotated with three veracity labels: support, not enough information (NEI), and contradict. 
As these datasets do not distinguish between partial and strong labels, we map Med-V1's prediction into the same three-way taxonomy by merging strong and partial support into support, strong and partial contradiction into contradict, and neutral to NEI.

\begin{figure}[h!]
    \centering
    \includegraphics[width=0.9\linewidth]{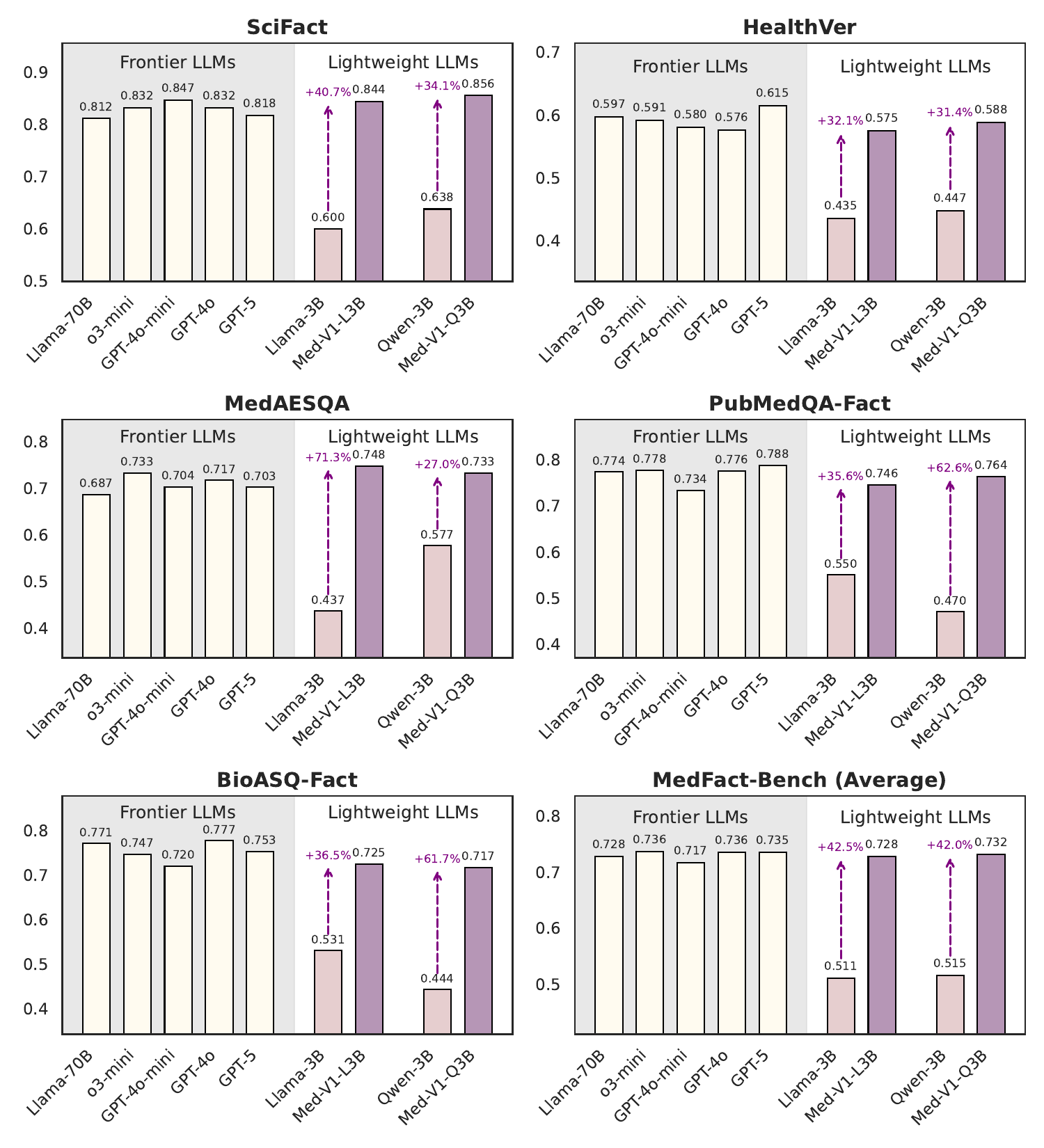}
    \caption{\textbf{Zero-shot accuracies of different LLMs on MedFact-Bench.} Performance is reported for each component dataset, and the (macro-)average accuracy is the main evaluation metric of MedFact-Bench. Frontier LLMs include large-scale open LLMs (e.g., 70B parameters) and the latest proprietary LLMs. Lightweight LLMs are 3B-parameter models. Med-V1-L3B is fine-tuned from Llama-3.2-3B-Instruct (Llama-3B), and Med-V1-Q3B is fine-tuned from Qwen2.5-3B-Instruct (Qwen-3B). Relative improvements over the respective base models are indicated. Both Med-V1 variants show significant performance improvements over their backbones. Llama-70B denotes Llama-3.3-70B-Instruct.}
    \label{fig:performance}
\end{figure}

Figure~\ref{fig:performance} reports the performance of different LLMs on MedFact-Bench. Notably, a wide variety of frontier LLMs, from Llama-3.3-70B-Instruct to GPT-4o and GPT-5, achieve closely clustered results across nearly all tasks, with average accuracies ranging from 0.717 to 0.736. We interpret this pattern as performance convergence on the current MedFact-Bench datasets rather than evidence that biomedical evidence attribution as a general task is saturated. The limited separation among models may partly reflect benchmark characteristics, including ambiguous claims, label noise, and artifacts introduced when binary question answering datasets are reformulated for evidence attribution, while the remaining knowledge and reasoning errors indicate continued room for model improvement.
In contrast, much smaller 3B-parameter models such as Llama-3.2-3B-Instruct and Qwen2.5-3B-Instruct perform substantially worse in this setting, with average accuracies of only 0.511 and 0.515, respectively. These gaps highlight the challenge of achieving reliable verification with lightweight models in the absence of domain-specific fine-tuning.

With training on MedFact-Synth, Med-V1 significantly improves over the base 3B-parameter models. On average, Med-V1-L3B achieves 42.5\% higher accuracy than Llama-3B (0.728 vs. 0.511), and Med-V1-Q3B achieves 42.0\% higher accuracy than Qwen-3B (0.732 vs. 0.515). These gains are consistent across all datasets and for both the Llama- and Qwen-based variants (27.0\% to 71.3\% relative performance improvements), demonstrating that the training pipeline is effective regardless of the underlying architecture. Overall, these results show that Med-V1 substantially narrows the performance gap between lightweight and frontier LLMs on biomedical evidence attribution, even though it is trained entirely on synthetic data and evaluated strictly in a zero-shot setting.

To quantify the contribution of GRPO reinforcement learning, we compare the SFT-only checkpoints with the final SFT+GRPO models using the same zero-shot MedFact-Bench evaluation. For Med-V1-L3B, GRPO increases the macro-average accuracy from 0.7198 to 0.7276 (+0.77 percentage points), with improvements on all five component datasets. For Med-V1-Q3B, the macro-average increases from 0.7294 to 0.7316 (+0.22 percentage points), although performance improves on SciFact and PubMedQA-Fact and decreases on HealthVer, MedAESQA, and BioASQ-Fact. These results indicate that SFT accounts for most of the final performance, while GRPO provides a modest, model- and dataset-dependent refinement (Supplementary Table~2).

To examine whether article-level overlap contributes to performance, we first compare the assertions in MedFact-Bench with the synthetic claims in MedFact-Synth. No MedFact-Bench assertion exactly matches a synthetic claim in MedFact-Synth based on either raw or normalized text. Among 6,868 resolved unique source PMIDs across MedFact-Bench, 558 (8.125\%) appear in MedFact-Synth as either claim-origin or verification-source articles. After evaluation instances associated with overlapping articles are excluded, the five-dataset macro-average changes from 0.7276 to 0.7266 for Med-V1-L3B and from 0.7316 to 0.7261 for Med-V1-Q3B. These small changes show that excluding overlapping articles had little effect on aggregate MedFact-Bench performance. Complete per-dataset overlap counts, retained sample sizes, and non-overlap accuracies are provided in Supplementary Table~3.

\subsection{Error Analysis Reveals High-quality Reasoning Despite Dataset Noise}
To better understand Med-V1's errors, especially its weaker performance on certain datasets, we conduct an error analysis across the five MedFact-Bench components. For each model variant (Med-V1-L3B and Med-V1-Q3B), we randomly sample 20 misclassified predictions from each dataset (100 total examples per model variant). Two annotators independently examine all 200 sampled cases, including the source article, claim, ground-truth label, and model output, and then discuss disagreements to reach a consensus label for each case. All reported error-analysis counts reflect these consensus-adjudicated labels. Based on this analysis, we categorize errors into four types: model prediction also acceptable (Type A), bad claim (Type B), model knowledge error (Type C), and model reasoning error (Type D).

Type A and B errors do not reflect true model failures. Type A errors consist of cases where, upon close inspection, the model's reasoning and answer are acceptable. For these cases, the resulting discrepancies between the ground truth and model error occurs either because the provided ground truth label does not accurately represent the claim-source pair, or because the model weighs evidence appropriately to conclude partial agreement or disagreement when the ground truth is neutral. The latter scenario occurs most frequently in the MedAESQA dataset. For instance, one MedAESQA claim states that ``Trauma-focused cognitive behavioral therapy (TFCBT), prolonged exposure therapy, eye movement desensitization and reprocessing (EMDR), and stress management are effective in reducing PTSD symptoms.'' While the source addresses the efficacy of EMDR and TFCBT, it does not mention prolonged exposure therapy or stress management. The model labels this claim-source pair as partial agreement, whereas the ground truth is neutral. However, because the source supports a subset of the information in the claim, the model’s classification as partial agreement is defensible.  Type B errors address claims that are of poor quality. For example, the claim is ambiguously phrased, not a true standalone assertion, or contains fabricated or non-interpretable biomedical concepts. Overall, Type B errors are less prevalent than others, but the frequency varies by dataset. BioASQ-Fact and MedAESQA contain the largest numbers of Type A cases: 17 and 15 cases, respectively, for Med-V1-L3B, and 16 cases each for Med-V1-Q3B (Table~\ref{tab:error_analysis}). In total, Types A and B account for the majority of the examined errors (68\% of the Med-V1-L3B sample and 71\% of the Med-V1-Q3B sample), suggesting that benchmark ambiguity and label quality contributed substantially to the reviewed discrepancies. 

The remaining two categories represent true model errors. Type C (knowledge errors) refers to inaccurate or incomplete factual knowledge and assumptions, such as misinterpreting the source, overlooking key evidence, or introducing unsupported details during reasoning. These errors are most frequent in SciFact ($n = 4$ and $3$ for Med-V1-L3B and Med-V1-Q3B, respectively) and PubMedQA-Fact ($n = 3$ for both variants). In contrast, Type D (reasoning errors) represents cases where the model correctly interprets the evidence but derives a self-contradictory or logically invalid conclusion. A notable pattern is that the step-by-step reasoning appears coherent, but ultimately reaches a verdict that contradicts its own evidence. Type D errors are most frequent in PubMedQA-Fact ($n = 8$) and SciFact ($n = 6$) for Med-V1-L3B, and in SciFact ($n = 6$) and HealthVer ($n = 5$) for Med-V1-Q3B.

The examined discrepancies therefore reflect several distinct limitations. Some arise from ambiguous claims, disputable ground-truth labels, or artifacts introduced when binary question answering examples are reformulated as claim verification instances. For example, the MedAESQA claim concerning multiple treatments for post-traumatic stress disorder is only partially supported by its source, making the model's partial-agreement prediction defensible despite a neutral ground-truth label. Other cases reflect genuine model limitations, including the knowledge error involving non-HDL cholesterol and the reasoning error involving febrile seizures. Because the same number of errors was sampled from each component dataset, the Type A--D proportions characterize the reviewed sample rather than the full benchmark-wide error distribution. Thus, convergence in aggregate accuracy across models should not be interpreted as a ceiling on biomedical evidence attribution. Further progress will require both improved models and more discriminative, carefully curated benchmarks.

\begin{table}[h!]
\centering
\caption{\textbf{Distribution of different error types for Med-V1 across MedFact-Bench.} Two annotators independently reviewed each sampled case and discussed disagreements to reach a final consensus label; the table reports these consensus-adjudicated counts. SF: SciFact; HV: HealthVer; MQA: MedAESQA; PQA: PubMedQA-Fact; BA: BioASQ-Fact. Type A: model prediction also acceptable; Type B: bad claim; Type C: model knowledge error; Type D: model reasoning error. Only Types C and D errors reflect true model failures.}
\label{tab:error_analysis}
\begin{tabular}{l*{12}{c}}
\toprule
\multirow{2}{*}{Error} 
    & \multicolumn{6}{c}{Med-V1-L3B} 
    & \multicolumn{6}{c}{Med-V1-Q3B} \\
\cmidrule(lr){2-7} \cmidrule(lr){8-13}
& SF & HV & MQA & PQA & BA & Sum 
& SF & HV & MQA & PQA & BA & Sum \\
\midrule
Type A  &  10   &  13  &  15  &  7  &  17  &  62  &  11  &  9  &  16 &  13  &  16  &  65 \\
Type B  &  0   &  3  &  1  &  2  &  0  &  6  &  0  &  5  &  1  &  0  &  0  &  6 \\
Sum     &  10   &  16  &  16  &  9  &  17  &  68  &  11  &  14  &  17  &  13  &  16  &  71 \\
\midrule
Type C  &  4   &  0  &  1  &  3  &  2  &  10  &  3  &  1  &  0  &  3  &  2  &  9 \\
Type D  &  6   &  4  &  3  &  8  &  1  &  22 &  6  &  5  &  3  &  4 & 2 & 20 \\
Sum     &  10   &  4  &  4  &  11  &  3  &  32  &  9  &  6  &  3  &  7  &  4  &  29 \\
\bottomrule
\end{tabular}
\end{table}

\begin{figure}[h!]
    \centering
    \includegraphics[width=\linewidth]{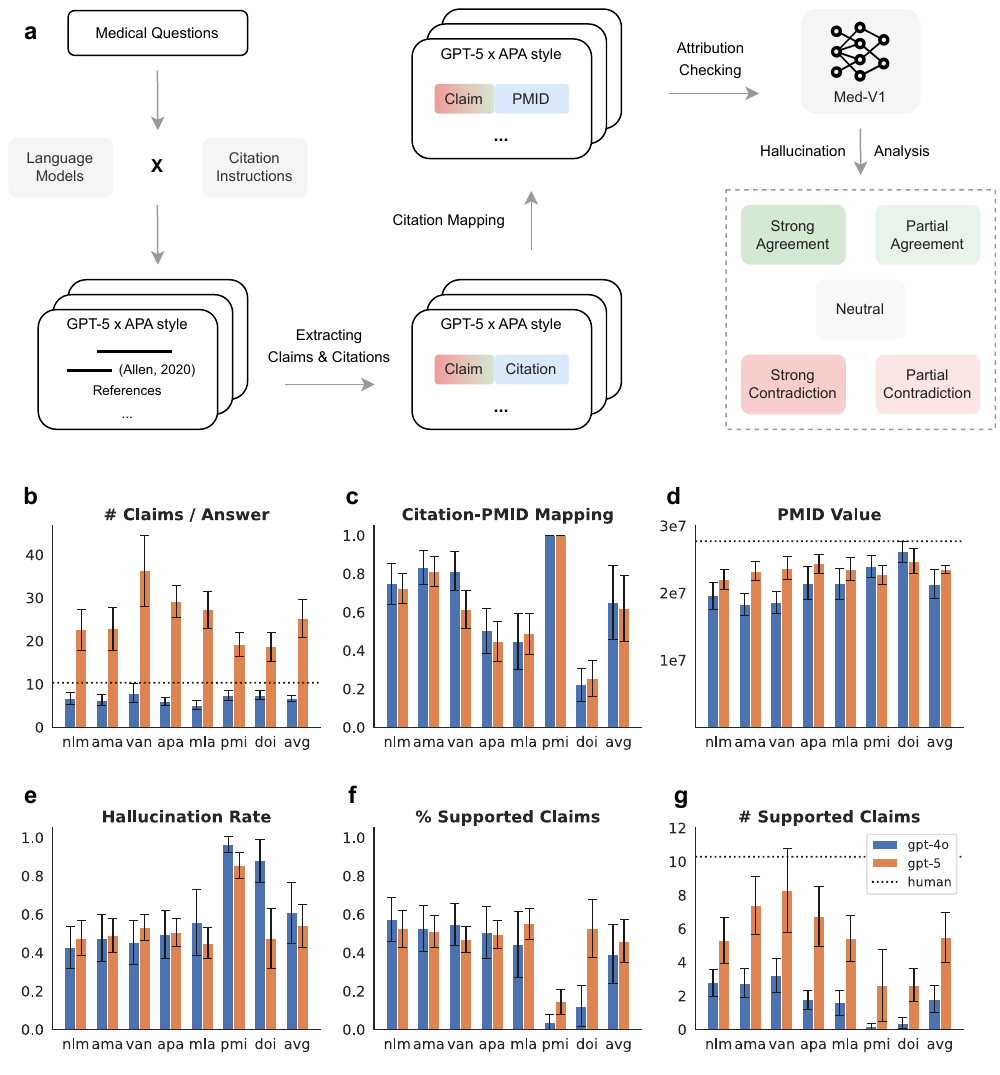}
    \caption{\textbf{Detecting LLM Hallucination with Med-V1}.
    \textbf{a}: Overview of this use case study. We use Med-V1 to analyze the hallucination rates of different LLMs and citation instructions. \textbf{b}: average number of claims (citation statements) per LLM-generated answer. \textbf{c}: The proportions of the generated citations that can be mapped to a PubMed ID (PMID). \textbf{d}: The average PMID values, which reflect their recency, generated by different methods. \textbf{e}: The hallucination rates of different methods. \textbf{f}: The proportion of supported claims generated by different methods. \textbf{g}: The number of supported claims generated by different methods. Statistics of the human-generated answers are shown in dotted horizontal lines. Avg: the averaged metric for each model. Error bars reflect 95\% confidence intervals estimated from 2,000 bootstrap iterations.}
    \label{fig:hallucination}
\end{figure}

\subsection{Use Case 1: Detecting LLM Hallucinations with Med-V1}
At a high level, this use case evaluates two distinct types of hallucination in citation-grounded LLM answers. Citation fabrication occurs when a generated citation cannot be mapped to a PubMed article in our pipeline. Content hallucination occurs when a citation maps successfully, but the mapped article does not partially or fully support the associated claim. Citation mapping therefore assesses citation fabrication, whereas Med-V1 assesses content hallucination among successfully mapped claim-article pairs.

We use MedAESQA because it provides medical questions together with human expert answers and citation-level evidence assessments linking individual answer statements to PubMed articles. This allows us to generate GPT-4o and GPT-5 answers to the same questions, compare their claims and supported claims with those in the human answers, and evaluate citation support at the individual claim-citation level. On MedAESQA, both Med-V1 variants outperform GPT-5 in citation-attribution accuracy (Med-V1-L3B: 0.748, Med-V1-Q3B: 0.733, GPT-5: 0.703; Figure~\ref{fig:performance}). We therefore use the first 40 MedAESQA questions to analyze citation fabrication and content hallucination in GPT-4o- and GPT-5-generated answers under seven citation instructions.

As shown in Figure~\ref{fig:hallucination}a, we prompt GPT-4o and GPT-5 using NLM (National Library of Medicine), AMA (American Medical Association), Vancouver, APA (American Psychological Association), MLA (Modern Language Association), and direct PMID or DOI citation instructions. Claim-citation pairs are extracted from each generated answer using GPT-5, and the citation texts are mapped to PMIDs using the PubMed single citation matcher \cite{fiorini2018user}. Titles and abstracts of the mapped PMIDs are retrieved from PubMed and provided to Med-V1-L3B together with the associated claims.

Supplementary Table~5 provides one illustrative example each of citation fabrication, content hallucination, and a supported citation. The examples include a GPT-4o direct-DOI citation that did not map to a PMID, a GPT-5 claim about emollients whose cited PMID mapped to an unrelated laparoscopic-imaging article and received a Med-V1 score of 0, and a GPT-4o side-sleeping claim whose AMA-formatted citation mapped to a relevant sleep-apnea article and received a score of +2.

On average, human experts answer with 10.3 claims for each question (Figure~\ref{fig:hallucination}b). In comparison, GPT-4o generates fewer claims (5.1 to 7.4) while GPT-5 generates many more claims (18.6 to 36.3). While GPT-4o generates similar numbers of claims under different citation styles, GPT-5 generates the most claims when using the Vancouver style. Overall, citations generated by GPT-4o and GPT-5 have similar rates of successful PMID mapping (Figure~\ref{fig:hallucination}c). Because we identify citation fabrication through mapping failure, lower mapping rates correspond to higher citation-fabrication rates. Except for directly citing PMIDs that gets perfect mapping, NLM, AMA, and Vancouver-style citations (61.4\% to 83.3\%) are more likely to be mapped successfully to PMIDs than APA and MLA citations (44.6\% to 50.4\%). Most of the LLM-generated DOIs are invalid and cannot be mapped to any PMID. Overall, the successful mapping rate of GPT-4o-generated citations is similar to that of GPT-5. Additionally, since PMIDs are indexed in a temporal order, we use their absolute values as a proxy for publication recency (Figure~\ref{fig:hallucination}d). Human-generated citations are the most up-to-date, with an average PMID value of 27696376, which corresponds to an article published in 2016. 

Among citations with successful PMID mapping, we define a content hallucination as a claim that is not partially or fully supported by the mapped PubMed source. Claims assigned Med-V1 scores of +1 (partial agreement) or +2 (strong agreement) are considered supported, whereas claims assigned scores of $-2$, $-1$, or 0 are considered content hallucinations. Among NLM, AMA, Vancouver, APA, and MLA citation styles, similar content hallucination rates are observed between GPT-4o (42.8\% to 55.8\%) and GPT-5 (44.9\% to 53.0\%), as shown in Figure~\ref{fig:hallucination}e. As expected, both models show high content hallucination rates (96.3\% for GPT-4o and 85.7\% for GPT-5) when instructed to directly cite PMIDs. Interestingly, while GPT-4o also shows a content hallucination rate above 80\% when citing DOIs, GPT-5 shows a lower content hallucination rate (47.5\%), which is consistent with other non-ID-based citation formats. This suggests that GPT-5 has improved memorization of DOI compared to GPT-4o. Figure~\ref{fig:hallucination}f shows the proportion of supported claims, which inversely reflects these discussed trends in hallucination rates.
Finally, while GPT-5 still generates more claims (2.6 to 8.3) that can be supported by the source than GPT-4o (0.2 to 3.2), both remain below the human baseline of 10.3 claims (Figure~\ref{fig:hallucination}g). For both models, the highest number of supported claims occurs under the Vancouver citation format, and the fewest occur when directly citing PMIDs or DOIs.

In summary, GPT-5 produces far more claims and citations than GPT-4o (especially under the Vancouver style), but both models show similar PMID-mapping patterns across citation styles and generally rely on less recent sources than humans. Among successfully mapped citations, their content hallucination rates are comparable for standard formats, whereas direct PMID citations yield very high content hallucination rates and DOI citations remain problematic for GPT-4o. Overall, both models still generate fewer supported claims than human experts.

\subsection{Use Case 2: Identifying High-Stakes Misattributions with Med-V1}
To illustrate the practical value of Med-V1 in real-world settings, we apply it at scale to clinical practice guidelines, which depend heavily on accurate citation and might impact high-stakes medical decision-making. We collect all publicly available guideline articles published between 2015 and 2025 in PubMed Central, resulting in 6,152 articles. From these, we extract decontextualized statements associated with a single citation, resulting in a total of 57k statement-source pairs. 

\begin{figure}[htbp]
    \centering
    \includegraphics[width=\linewidth]{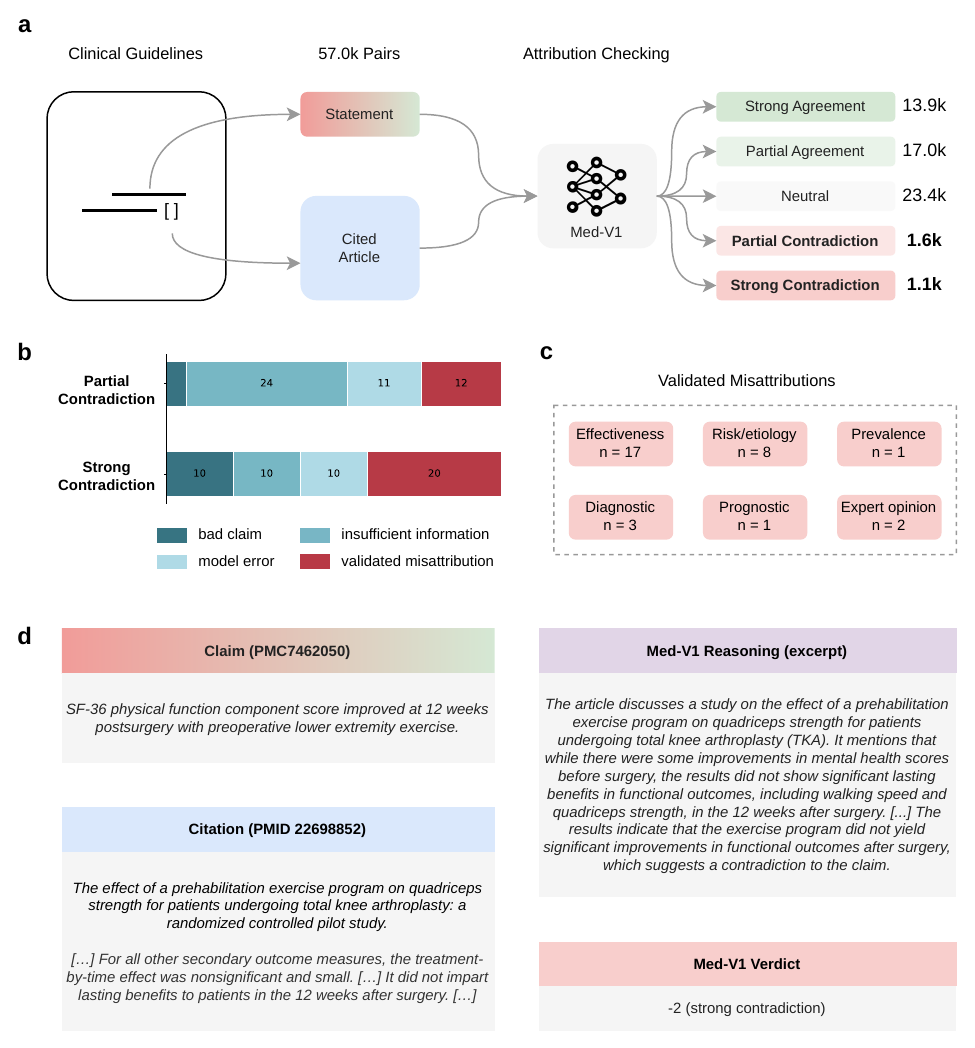}
    \caption{\textbf{Identifying High-stakes Misattributions with Med-V1}.
    \textbf{a}: Overview of this use case study. We extract citation statements and their source articles from clinical guidelines and automatically check their attribution validity with Med-V1. \textbf{b}: Distribution of the consensus-adjudicated manual review of 50 partial contradiction and 50 strong contradiction samples. \textbf{c}: Clinical-domain distribution of the 32 consensus-validated misattributions. \textbf{d}: Example of a consensus-validated treatment-effectiveness misattribution. The guideline attributes improved SF-36 physical function at 12 weeks after total knee arthroplasty to preoperative lower-extremity exercise, whereas the cited randomized pilot study reports nonsignificant treatment-by-time effects for the other secondary outcomes and no lasting benefit during the 12 weeks after surgery. Med-V1 assigns a score of $-2$ (strong contradiction).}
    \label{fig:misattribution_fig}
\end{figure}

For each pair, we apply Med-V1-L3B to classify if the cited article supports the citation statement. As shown in Figure~\ref{fig:misattribution_fig}a, most citations (54.2\%) are predicted to be strongly (24.4\%) or partially (29.8\%) supported by the source, which is expected as guideline recommendations are typically grounded in evidence that the authors regard as supportive.  
Importantly, a smaller portion is flagged as partial contradiction (1.6k pairs, 3\%) or strong contradiction (1.1k pairs, 2\%). In such cases, a guideline statement might cite a source that does not support the claim or even directly contradicts it. To assess the validity of these flagged cases, we randomly sample 100 contradictions for review, including 50 partial contradiction and 50 strong contradiction cases. Two annotators independently evaluate all 100 cases and assign each case to one of five categories: validated misattribution, poor quality claim, not enough information arising from the use of titles and abstracts as the source instead of full text, model error, or other explanation. They then discuss disagreements to reach a final consensus category. All reported validation and domain counts reflect these consensus-adjudicated labels. 

After consensus adjudication, 32 cases are validated as misattributions, meaning that the citation statement in the guideline is contradicted by the cited article (Figure~\ref{fig:misattribution_fig}b).
To characterize their potential impact, we categorize claims into six domains: effectiveness, risk/etiology, prevalence, diagnostic, prognostic, and expert opinion/policy. The most frequently misattributed domain is treatment effectiveness, accounting for 17 of 32 validated misattributions (53.1\%; Figure~\ref{fig:misattribution_fig}c). For example, one guideline states that the ``SF-36 physical function component score improved at 12 weeks postsurgery with preoperative lower extremity exercise.'' The cited randomized pilot study instead reports that treatment-by-time effects for the other secondary outcomes were nonsignificant and small and that the intervention did not provide lasting benefits during the 12 weeks after surgery. This citation therefore does not support the specific attributed improvement in SF-36 physical function at 12 weeks (Figure~\ref{fig:misattribution_fig}d). The second most common category is risk/etiology, accounting for 8 of 32 validated misattributions (25.0\%). For instance, one claim states ``There are several compelling evidences that childhood OB is the most common risk factor for both cardiovascular (CV) and metabolic diseases,'' while the corresponding citation mentions that when body mass index (BMI) is adjusted for, there is a weak negative association between childhood BMI and metabolic variables. This example highlights how misattributions can arise from misrepresenting association findings. These two categories of misattributions are particularly consequential, as they can directly inform decisions regarding treatment and prevention. 

It is important to highlight that a claim categorized as a validated misattribution does not necessarily indicate that the claim itself is universally incorrect, rather that it is not supported by its cited source. Overall, this large-scale analysis demonstrates that Med-V1 can efficiently audit citation accuracy in biomedical documents with potential public health impact. This approach provides a practical way to identify inconsistencies that would be extremely difficult to detect manually at the guideline scale.

\section{Discussion}

In this work, we introduce Med-V1, a family of 3B-parameter models for biomedical evidence attribution trained on the large-scale MedFact-Synth corpus to produce both structured verdicts and natural-language explanations. Across MedFact-Bench, Med-V1 substantially improves over lightweight base models and narrows the performance gap with much larger frontier LLMs, while remaining far more efficient to deploy. Analyses of MedFact-Synth indicate that the synthetic supervision is diverse and consistent, and the error analysis suggests that a substantial fraction of the manually examined benchmark discrepancies reflect dataset ambiguity or label noise rather than systematic failures of reasoning. These results support the view that targeted post-training on high-quality synthetic supervision can unlock strong verification ability in lightweight models.

The similar performance on the non-overlap subsets supports generalization beyond the shared source articles.

Biomedical evidence attribution faces a persistent supervision bottleneck because high-quality claim-evidence labeling is expensive and difficult to scale across medical subdomains. Prior work on biomedical and scientific claim verification has largely used supervised encoder or encoder-decoder architectures organized as pipeline or joint systems for document retrieval, evidence selection, and veracity prediction. SciFact established this formulation using expert-written scientific claims with document-level labels and sentence-level rationales \cite{wadden2020fact}. VerT5erini uses a T5-based pipeline for abstract retrieval, sentence selection, and label prediction \cite{pradeep-etal-2021-scientific}; ARSJoint jointly models abstract retrieval, rationale selection, and stance prediction \cite{zhang-etal-2021-abstract}; and MultiVerS predicts document-level labels and extractive rationales from full-document context \cite{wadden-etal-2022-multivers}. HealthVer and HealthFC extend claim verification to real-world health claims, with HealthFC evaluating pipeline and joint systems based on biomedical and general-domain encoders \cite{sarrouti2021evidence,vladika-etal-2024-healthfc}. Open-domain work has further shown that knowledge-source and retrieval choices affect downstream verification \cite{vladika-matthes-2024-comparing}, while more recent studies have explored iterative LLM-based reasoning for medical claims \cite{vladika-etal-2025-step}. These approaches provide strong task-specific systems and, in many cases, identify extractive evidence rationales.

Med-V1 differs from these approaches in four practical respects. First, it is a lightweight 3B-parameter model that is substantially more efficient to deploy at scale than frontier LLMs. Second, it is evaluated in a zero-shot setting across heterogeneous claim verification, citation attribution, and repurposed binary question answering tasks rather than being fine-tuned separately for each benchmark. Third, Med-V1 produces both a structured five-point verdict and a natural-language explanation, whereas earlier systems commonly produce classification labels, extractive evidence rationales, or both. Finally, the Med-V1 model checkpoints, MedFact-Synth training data, MedFact-Bench evaluation resources, and training code are publicly available. Because prior studies differ in their datasets, evidence inputs, retrieval assumptions, supervision regimes, and evaluation metrics, we compare these approaches descriptively rather than drawing conclusions from cross-paper performance values.

Recent work on synthetic verification data \cite{wright-etal-2022-generating,zhang2025enhancinghealthfactcheckingllmgenerated} has sought to reduce reliance on manually annotated examples by generating supported and refuted claims or other synthetic supervision. Med-V1 builds on this direction by pairing generated claims with independently retrieved biomedical articles, assigning five-point labels through multi-model consensus, and supervising both structured verdicts and natural-language explanations.

The two use cases illustrate how a lightweight verifier can serve as reusable infrastructure for auditing. The hallucination analysis shows that citation instructions strongly influence citation mappability and downstream support rates, implying that citation format is not merely a presentation choice but can alter factual reliability. The guideline audit shows that automated verification can surface potentially high-stakes misattributions at a scale that is otherwise infeasible, enabling risk-based triage in which domain experts review a focused set of flagged contradictions. These applications do not require perfect verification accuracy to be valuable, because operational utility often depends on high precision among the most suspicious cases rather than uniform performance across all instances.

This study has several limitations. First, Med-V1 was trained and primarily evaluated on standalone claims paired with PubMed titles and abstracts. Extending Med-V1 to full-text evidence attribution is an important future direction, but foundational infrastructure is still needed, including benchmark datasets with claim-level labels, evidence-span provenance, and measures of long-context generation stability. Second, we evaluated Med-V1 as a verifier rather than as part of an end-to-end retrieval-and-verification system, so practical performance will also depend on evidence retrieval. Third, the current formulation evaluates individual claim-article relationships and does not directly assess evidence quality, study design, generalizability, or risk of bias. Fourth, the human evaluation included 100 consensus-filtered MedFact-Synth instances and therefore provides an initial rather than comprehensive assessment of label quality. The consensus procedure also does not assign a separate confidence score, and the available analysis does not establish that the aggregation rule is optimal. Fifth, we did not train separate models using individual-model labels or label-only targets, so the independent contribution of these design choices remains unquantified. Finally, the close performance of several models on MedFact-Bench may partly reflect ambiguous claims, label noise, and artifacts introduced when binary question answering datasets are reformulated for evidence attribution. Similar performance plateaus on PubMedQA and other established biology benchmarks have been reported despite continued gains on more challenging evaluations \cite{justen2025llms}. The remaining knowledge and reasoning errors also indicate continued opportunities for model and benchmark improvement.

Taken together, our findings suggest that high-quality synthetic supervision, paired with targeted post-training, can serve as a practical alternative to expensive frontier LLMs for many biomedical evidence-attribution tasks. Beyond Med-V1 itself, our framework for generating reasoning-aware synthetic corpora and using them to align small open models may generalize to other domains that require transparent and scalable verification, such as clinical document summarization. As the volume of AI-generated biomedical text continues to grow, lightweight evidence-attribution models like Med-V1 may play an increasingly important role as reusable infrastructure for auditing the factual consistency and citation integrity of automated systems.

\section{Methods}\label{methods}

\subsection{Constructing MedFact-Synth with synthetic data generation}

\begin{figure}[htbp]
    \centering
    \includegraphics[width=0.5\linewidth]{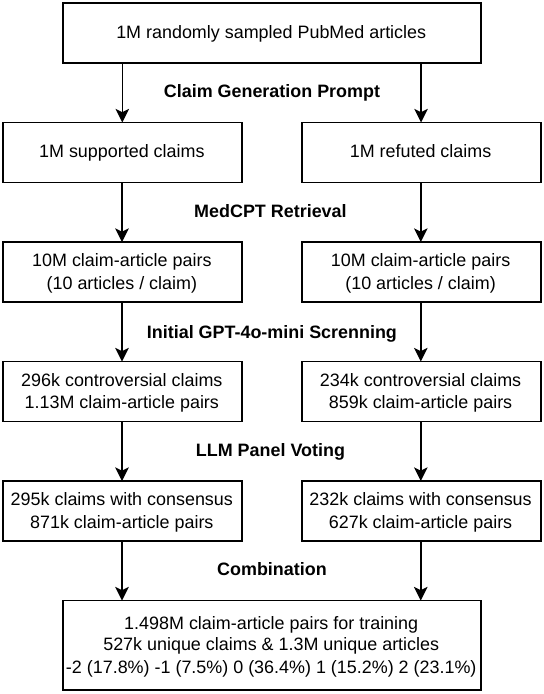}
    \vspace{1cm}
    \caption{\textbf{Overview of the proposed pipeline for synthetic data generation.} One million PubMed articles are randomly sampled and used to generate two standalone claims per article: one potentially supported and one potentially refuted by the source article. Each claim is paired with ten candidate articles retrieved using MedCPT, resulting in approximately 20 million claim-article pairs. An initial screening stage using GPT-4o-mini filters for controversial claim-article pairs that receive mixed verdict signals. These pairs are then evaluated by a panel of frontier LLMs, and a consensus label is assigned when sufficient agreement is reached. Finally, supported- and refuted-claim pipelines are merged to construct MedFact-Synth, yielding 1.5 million high-quality claim-article pairs with five-point Likert-scale veracity labels and natural-language rationales, covering a balanced distribution of agreement, neutrality, and contradiction.}
    \label{fig:pipeline}
\end{figure}

Figure \ref{fig:pipeline} summarizes the pipeline used to construct MedFact-Synth, our large-scale synthetic corpus for supervising biomedical evidence attribution. The pipeline operates in three main stages: (i) sampling biomedical articles and generating synthetic claims from them, (ii) using dense retrieval to match each claim to potentially relevant articles, and (iii) assigning consensus veracity labels and rationales using a panel of LLMs. 

\textbf{Stage 1: Sampling articles and generating claims.}
We begin from a local copy of the PubMed 2025 baseline and randomly sample one million articles with non-empty titles and abstracts. For each sampled article, we prompt GPT-4o-mini twice: once to generate a claim that can be supported by the article (by Prompt~\ref{prompt:supported}), and once to generate a claim that can be refuted by it (by Prompt~\ref{prompt:refuted}). In both cases, the model is instructed to produce standalone, interpretable statements that do not rely on the original text for context. For refuted claims, we additionally discourage trivial negations (e.g., simply inserting ``not'' into an otherwise supported claim). This procedure yields two synthetic claims per article, each tied to the source paper but phrased for independent verification. 

\begin{AIbox}{Supported Claim Generation Prompt}
\label{prompt:supported}
\textbf{System Prompt}\\
You are a biomedical annotation expert, and your task is to generate a claim that can be supported by the provided article. The claim should be interpretable on its own, without relying on the article. Do not generate anything else than the claim.
\\\\
\textbf{User Prompt}\\
Here is the article:\\
Title: \{title\}\\
Abstract: \{abstract\}
\end{AIbox}

\begin{AIbox}{Refuted Claim Generation Prompt}
\label{prompt:refuted}
\textbf{System Prompt}\\
You are a biomedical annotation expert, and your task is to generate a claim that can be refuted by the provided article. The claim should be interpretable by its own, without relying on the article. Avoid using simple negative words such as not and no. Do not generate anything else than the claim.
\\\\
\textbf{User Prompt}\\
Here is the article:\\
Title: \{title\}\\
Abstract: \{abstract\}
\end{AIbox}

\textbf{Stage 2: Matching claims to candidate evidence.}
Rather than always pairing each claim with its source article, we aim to mimic realistic verification settings in which a claim must be checked against independently retrieved literature. To achieve this, we use a dense retrieval system based on MedCPT \cite{jin2023medcpt} to encode each synthetic claim into an embedding and search a large index of PubMed article embeddings. For every claim, we retrieve the top $k$ most similar articles (with $k=10$ in our experiments), together with their titles and abstracts. This step decouples claims from their originating papers and naturally produces a diverse set of claim-article pairs spanning strong agreement, partial agreement, neutrality, and contradiction.

\textbf{Stage 3: Multi-LLM reasoning and consensus labeling.}
In the final stage, we assign high-quality veracity labels and rationales to the retrieved claim-article pairs using a panel of LLMs. Each pair is first evaluated by GPT-4o-mini with the Med-V1 Verification Prompt (Prompt~\ref{prompt:verification}), which instructs the model to (i) reason step by step about how the article relates to the claim, and (ii) assign a discrete score on a five-point Likert scale: strong contradiction ($-2$), partial contradiction ($-1$), neutral/unrelated ($0$), partial agreement ($+1$), or strong agreement ($+2$). This initial pass yields both scores and rationales for tens of millions of claim-article pairs.

To focus labeling resources on the most informative examples, we then identify ``controversial'' claims whose retrieved articles receive a mixture of supportive, contradicting, and neutral scores. For these challenging pairs, we collect additional evaluations from two stronger LLMs (Llama-3.3-70B-Instruct and o3-mini), resulting in three independent scores and associated rationales per pair. We derive a consensus label when at least two of the three models assign the same score and the third model differs from that score by no more than one point. The repeated score is retained as the consensus label. Thus, a panel of $(+1,+1,0)$ is assigned $+1$, indicating partial agreement, whereas a panel of $(+2,+2,+1)$ is assigned $+2$, indicating strong agreement. The rule therefore preserves differences in attribution strength, although both panels satisfy the same inclusion criterion. For pairs that meet this criterion, we randomly select one rationale from one of the agreeing models and store it alongside the final label, the claim, and the article abstract.

Applying these filters produces MedFact-Synth, a high-quality synthetic dataset containing approximately 1.5 million claim-article pairs, covering hundreds of thousands of unique claims and over one million unique articles. Each instance includes (i) a standalone biomedical claim, (ii) a title-abstract pair serving as the verification source, (iii) a structured veracity label on the five-point Likert scale, and (iv) a natural language explanation generated by the LLM panel. This reasoning-aware supervision is subsequently used to post-train Med-V1.

MedFact-Synth was developed over approximately four months, from May to August 2025. This elapsed period reflects the end-to-end development process rather than active compute time and includes API rate limits and service constraints, asynchronous and overlapping jobs, retries, intermediate quality-control checks, and pauses between pipeline stages. The final MedFact-Synth dataset is publicly available, enabling researchers to use the resulting resource without repeating the full construction process.

\begin{AIbox}{Med-V1 Verification Prompt}
\label{prompt:verification}
\textbf{System Prompt}\\
You are a fact-checking expert trained in evidence-based medicine. Your task is to evaluate how strongly an *article* agrees or disagrees with a *claim*. The *article* is retrieved from a search engine using the *claim* as the query.\\
Use the following five-point scale:\\
   - **-2 Strong Contradiction**  – The article clearly and directly refutes the claim.\\
   - **-1 Partial Contradiction** – The article provides mixed or indirect evidence against the claim.\\
   - ** 0 Neutral / Unrelated**   – The article does not address the claim, offers insufficient information, or is irrelevant to the claim.\\
   - ** 1 Partial Agreement**	 – The article offers some indirect or tentative support for the claim.\\
   - ** 2 Strong Agreement**	 – The article explicitly and strongly supports the claim.\\
Note that the *article* might not describe the exact same subjects, interventions, or measurements as the *claim*. In this case, please note the difference and assign a score of 0. \\
Output in two parts only and do not output anything else:\\
\textless think\textgreater[your detailed, step‐by‐step explanation for scoring]\textless /think\textgreater\\
\textless score\textgreater[the integer score only, i.e., -2, -1, 0, 1, or 2]\textless /score\textgreater
\\\\
\textbf{User Prompt}\\
Article:\\
\{source-article\}\\ \\
Claim:\\
\{claim\}
\end{AIbox}

\subsection{Manual validation of MedFact-Synth}

To assess the quality of the synthetic labels, we randomly sample 100 instances from MedFact-Synth. Three annotators (Z.W., N.W., J.C.) with biomedical training independently evaluate each claim-source pair using the five-point label definitions in the Med-V1 Verification Prompt (Prompt~\ref{prompt:verification}). A score of $-2$ indicates strong contradiction, in which the article clearly and directly refutes the claim; $-1$ indicates partial contradiction, in which the article provides mixed or indirect evidence against the claim; $0$ indicates that the article does not address the claim, provides insufficient information, or is irrelevant; $+1$ indicates partial agreement, in which the article provides indirect or tentative support; and $+2$ indicates strong agreement, in which the article explicitly and strongly supports the claim. When the article concerns different subjects, interventions, or measurements from the claim, annotators assign a score of $0$. Annotators base their judgments solely on the title and abstract of the source article. After the independent annotation round, disagreements are discussed to establish a human consensus label for each instance.

We evaluate exact five-point agreement and three-category agreement with the human consensus labels. Exact five-point agreement requires an identical score. For three-category agreement, we group scores of $-2$ and $-1$ as negative, 0 as neutral, and $+1$ and $+2$ as positive. The single midpoint value initially recorded as $+1.5$ during adjudication was resolved to $+1$ so that all human consensus labels follow the predefined integer scale.

\subsection{Training Med-V1}

We use a two-stage post-training procedure that first applies supervised fine-tuning (SFT) and then reinforcement learning (RL) \cite{guo2025deepseek}. Using this framework, we train two variants of Med-V1: Med-V1-L3B, initialized from Llama-3.2-3B-Instruct \cite{grattafiori2024llama3herdmodels}, and Med-V1-Q3B, initialized from Qwen2.5-3B-Instruct \cite{qwen2025qwen25technicalreport}. Both variants follow the same training schedule and hyperparameters. Their only difference is the backbone model.

\paragraph{Supervised fine-tuning.}

In the SFT stage, we formulate verification as a standard conditional generation task. Given a claim and a source article, the model is trained to produce the corresponding reasoning trace and veracity score provided by MedFact-Synth. We supervise both the reasoning trace and veracity score because Med-V1 is intended to generate a natural-language explanation together with its structured verdict at inference; a label-only target would not directly supervise this explanation-generation capability. We optimize a language modeling objective using the AdamW optimizer with a learning rate of $2\times 10^{-5}$, an effective batch size of 128, and train for 10 epochs on approximately 1.5 million instances. All SFT experiments are conducted on Nvidia H200 (140GB) GPUs with distributed data-parallel training. In this setup, training Med-V1 on 8$\times$H200 GPUs requires on the order of tens of GPU-days.

\paragraph{Reinforcement learning.}

After SFT, we apply RL to further refine the model's alignment with human preferences. 
Specifically, we design a rule-based reward function that evaluates each prediction in two steps. First, it checks whether the output adheres to the format required in the Med-V1 verification: one \texttt{<think>...</think>} block followed by one \texttt{<score>...</score>} block. Any output that violates this format receives a reward of -1. 
For well-formatted predictions, the function then extracts the model's predicted score and compares it with the ground-truth label. Outputs with invalid or uninterpretable scores are again assigned a reward of -1. Valid scores are rewarded according to $0.5 \times \left(2 - \lvert y_{\text{pred}} - y_{\text{true}} \rvert \right)$, which assigns higher values to predictions whose 5-point Likert scores are closer to the ground truth.

We then update the LLMs (policy model) using Group Relative Policy Optimization (GRPO)~\cite{shao2024deepseekmath}, which estimates advantages without a separate critic model. For each prompt, the policy generates multiple candidate responses, and each is scored by the rule-based reward function. The mean reward of all responses for a given prompt serves as the baseline, and each response is compared with it to compute its advantage.
Similarly, we perform GRPO training on 1.5M MedFact-Synth instances.
For each prompt, the policy generates 5 rollouts, all evaluated by the rule-based reward function. The actor model is initialized from the SFT checkpoint and trained with a maximum prompt length of 3,072 tokens and a maximum response length of 1,024 tokens. We use AdamW with a learning rate of $1\times10^{-3}$ and a global batch size of 1,440. KL regularization follows the low-variance formulation with a coefficient $10^{-3}$ to constrain drift from the SFT model. 
Training runs for 3 epochs on a single node with 8$\times$Nvidia H200 (140GB) GPUs, for a total of 3,120 steps.

\subsection{Developing MedFact-Bench}

To evaluate Med-V1, we curate MedFact-Bench, which unifies five biomedical benchmarks spanning claim verification, citation attribution, and repurposed binary question answering in a common assertion-source evaluation format. The component datasets are SciFact \cite{wadden2020fact}, HealthVer \cite{sarrouti2021evidence}, MedAESQA \cite{gupta2025dataset}, PubMedQA-Fact (repurposed from PubMedQA \cite{jin2019pubmedqa}), and BioASQ-Fact (repurposed from BioASQ \cite{krithara2023bioasq}). Each instance in MedFact-Bench consists of a claim-source pair annotated with one of three labels: support, not enough information (NEI), or contradict. Because Med-V1 produces a 5-point Likert score that reflects the degree of agreement between the claim and the source, we map the model's predictions into this three-way taxonomy: strong/partial agreement $\rightarrow$ support, strong/partial contradiction $\rightarrow$ contradict, and neutral $\rightarrow$ NEI. We report macro-average accuracy across the five datasets as the primary metric, as it equally weights each dataset and reduces the influence of varying dataset sizes. Illustrative examples from each MedFact-Bench component dataset are provided in Supplementary Table~1, including the original task input, the assertion used for evaluation, the source article or identifier, the original dataset label, and the corresponding unified MedFact-Bench label.

\paragraph{Train-test article-overlap analysis.} To assess potential article-level overlap between training and evaluation, we compare the source PMIDs in each MedFact-Bench component with PMIDs appearing in MedFact-Synth as either claim-origin articles or verification-source articles. We classify an evaluation instance as overlapping when its uniquely resolved source PMID appears in either training role. Instances with unresolved or ambiguous source PMIDs are reported separately and are not assumed to be non-overlapping. We recalculate accuracy using only evaluation instances whose resolved source PMID does not occur in MedFact-Synth. We also compare raw and normalized assertion text for exact matches. Here, zero-shot indicates that Med-V1 was not fine-tuned on benchmark-specific labels or training examples; it does not imply complete article-level disjointness between MedFact-Synth and MedFact-Bench.

\paragraph{Claim verification datasets.}
SciFact \cite{wadden2020fact} and HealthVer \cite{sarrouti2021evidence} are established biomedical claim verification datasets that provide diverse claim-source pairs annotated with support, contradict, or NEI labels. SciFact contains expert-curated scientific claims derived from citation statements and paired with abstracts that include the relevant evidence. HealthVer comprises real-world medical claims verified against biomedical abstracts. For consistency, we use the pre-processed versions of these datasets \cite{wadden-etal-2022-multivers}, containing 340 SciFact instances and 903 HealthVer instances.

\paragraph{Citation attribution dataset.}
MedAESQA \cite{gupta2025dataset} is an expert-annotated benchmark designed to evaluate citation attribution in LLM-generated health answers. Each example contains an LLM-generated answer with one or more citation markers, and annotators label whether the cited PubMed article supports the associated statement. We preprocess MedAESQA by flattening each answer into individual statement-PMID pairs, removing bracketed citation markers from the statement text, and normalizing sentence formatting. We then map each cited PMID to the title and abstract of the corresponding PubMed record in our local PubMed 2025 baseline, and discard pairs with missing or invalid PMIDs. 
MedAESQA provides fine-grained labels including supporting, contradicting, neutral, and not relevant. To include MedAESQA in MedFact-Bench under a unified three-way claim verification taxonomy, we map neutral and not relevant into NEI. This yields a subset of 9,106 verification instances.

\paragraph{Re-purposed Boolean QA datasets.}
In addition to these established verification datasets, MedFact-Bench includes two re-purposed Boolean QA datasets, which we refer to as PubMedQA-Fact and BioASQ-Fact. We convert each QA instance into a verification example using a two-step process. First, each yes/no question is rewritten as a declarative claim under the assumption that the answer is ``yes.'' Second, we map the original QA answers to verification labels: yes $\rightarrow$ support, maybe $\rightarrow$ NEI, and no $\rightarrow$ contradict. This repurposing transforms large-scale QA datasets into fact-checking benchmarks compatible with the unified MedFact-Bench framework.
PubMedQA-Fact is derived from PubMedQA \cite{jin2019pubmedqa}, where each question corresponds to a PubMed title and the associated source is the abstract with its conclusion removed (the ``reasoning-required'' setting). In MedFact-Bench, we include the re-purposed PubMedQA test set of 500 instances. 
BioASQ-Fact is constructed from the BioASQ yes/no QA task \cite{krithara2023bioasq}, whose questions and answers are curated by biomedical experts and each supported by PubMed abstracts. In MedFact-Bench, we include all 3,425 yes/no questions from the BioASQ-13B training set.

\begin{AIbox}{Question Conversion Prompt}
\textbf{System Prompt}\\
You are a helpful assistant. Your task is to convert a yes/no question into a declarative statement. The statement should be a claim that is true if the answer to the question is ``yes''. Do not output anything else than the converted statement. \\\\
\textbf{User Prompt}\\
Convert the following question into a claim, assuming the answer is ``yes'': \\
Question: \{question\}
\end{AIbox}

\subsection{Error Analysis}
We conduct an error analysis using the Med-V1 predictions of MedFact-Bench. From each of the component datasets of MedFact-Bench, we randomly sample 20 cases in which Med-V1 produced a label different from the ground truth, resulting in 100 cases for in-depth evaluation for each model variant. Two annotators independently review every sampled case, including the claim, source abstract, ground-truth label, and Med-V1-generated reasoning. After completing their independent assessments, the annotators discuss disagreements and reach a consensus label for each case. All reported error-analysis counts reflect these consensus-adjudicated labels. Our objective is to understand and characterize the underlying reasons for the discrepancies between ground truth and model output. We document the source of error and derive a set of four error categories to classify these cases. We use a hierarchical procedure to ensure that only one error category is assigned to each case. 

We first assess the claim itself, and if it is of poor quality, then the case is immediately labeled as a Type B (bad claim) error, and no further analysis is conducted. An example is the ambiguous claim ``a popular treatment to tamp down the immune system in severely ill patients may help a few, but could harm many others,'' where ``popular treatment'', ``a few'', and ``others'' lack the necessary context to draw a conclusion. 

For cases with coherent claims, we review the source and draw a conclusion about the claim’s validity. We compare this conclusion to the ground truth label, and if the label is contradictory but the model's prediction matches our classification, we assign the case as Type A (model prediction also acceptable). For instance, the claim ``there is software for automated analysis of FISH images'' is labeled as supported in the ground truth, but the source is about creating FISH images and does not discuss image analysis techniques, so the correct label should be ``not enough information''. 

If the claim and ground truth are consistent with each other, then we next examine Med-V1’s reasoning to identify the source of error. For the special case in which the ground truth is neutral, but Med-V1 correctly identifies that some components of the claim are directly supported or refuted by the source while others are not mentioned, thereby concluding a partial agreement or disagreement, we assign the case Type A. For true model errors, we assign Type C (knowledge-based) or Type D (reasoning-based). For the claim ``Non-HDL-cholesterol is a better predictor of long-term outcome in patients after acute myocardial infarction compared to LDL-cholesterol,'' Med-V1 makes a knowledge error through misinterpreting the source's evidence regarding LDL-cholesterol, resulting in a contradictory label when the claim is actually supported. In contrast, a reasoning-based error arises for the claim ``Febrile seizures reduce the threshold for development of epilepsy,'' for which the model incorrectly concludes that the source supports the claim, even though it also acknowledges that the source does not directly confirm the claim. This method is applied across both Med-V1-L3B and Med-V1-Q3B outputs. We tabulate the error counts across a total of 200 cases and review the distribution of categories by type and dataset. 

\subsection{Detecting LLM Hallucinations with Med-V1}
We conduct a use case study to quantify citation fabrication and content hallucination in citation-grounded LLM answers under different citation instructions. We selected GPT-4o and GPT-5 because they were widely deployed frontier LLMs at the time of the study and represented successive model generations from the same provider, enabling us to compare their citation behavior using the same prompting and downstream evaluation pipeline. We use the first 40 medical questions from MedAESQA \cite{gupta2025dataset} and prompt the two models to generate answers under seven citation-instruction formats: NLM, AMA, Vancouver, APA, MLA, and directly citing PMIDs or DOIs. 

\begin{AIbox}{Claim Extraction Prompt}
\label{prompt:claim_extraction}
\textbf{System Prompt}\\
You are an expert in biomedical literature and citation analysis. Your task is to extract every factual claim and its corresponding full citation from the provided text.\\

Instructions:\\
1. Identify every sentence or clause that makes a factual claim supported by a citation.\\
2. For each claim, identify the inline citation marker (e.g., ``[1]'', ``(Smith, 2023)'', ``¹'', ``(PMID: 12345)'').\\
3. Resolve this inline citation to its \textbf{full reference entry} from the bibliography/reference list at the end of the text.\\
\hspace*{1em}-- If the text uses numeric citations (e.g., AMA, Vancouver, NLM), match the number to the numbered reference list.\\
\hspace*{1em}-- If the text uses author-date/page citations (e.g., APA, MLA), match the author/date or author/page to the alphabetical reference list.\\
\hspace*{1em}-- If the text uses only inline identifiers (e.g., PMID, DOI) and has no reference list, use the full inline citation string itself (e.g., ``PMID: 12345'').\\
4. Output the results as a strict JSON list of objects (\textbf{only} JSON; no additional text).\\
5. If a claim has multiple citations, repeat the same claim multiple times, once per citation, each time with a different citation.\\[0.2em]

JSON Format (example structure):\\
\texttt{[}\\
\hspace*{1em}\texttt{\{}\\
\hspace*{2em}\texttt{"claim": "The exact text of the factual claim.",}\\
\hspace*{2em}\texttt{"citation": "The full text of the corresponding reference entry (e.g., '1. Author AA. Title. Journal. Year...')."}\\
\hspace*{1em}\texttt{\}}\\
\texttt{]}\\[0.2em]

\textbf{User Prompt}\\
\{model\_answer\}
\end{AIbox}

To evaluate citation attribution, we apply a two-step extraction and processing workflow. First, GPT-5 is instructed with Prompt~\ref{prompt:claim_extraction} to extract all factual claims from each generated answer and to resolve each in-text citation marker to the corresponding full reference entry when a reference list is present. This produces a set of claim-citation pairs per answer. Second, we parse and normalize the extracted citations. For all styles, we remove leading numbering and trailing punctuation. For PMID- and DOI-based instructions, we further extract the bare identifier string. For non-identifier citation styles (e.g., NLM and AMA), we retain the full citation text after normalization.

We then map citations to PubMed articles. For non-identifier citation styles, we map each citation string to PMIDs using the PubMed single citation matcher \cite{fiorini2018user}, retaining the top-ranked match returned by the system. For DOI-based citations, we resolve DOIs to PMIDs using the National Center for Biotechnology Information (NCBI) ID Converter API \cite{sayers2021database} and standardize them into PMID-based mappings. Generated citations that cannot be mapped to a PubMed article are treated as citation fabrications. For successfully mapped PMIDs, we retrieve titles and abstracts from a local PubMed 2025 baseline and use them as the verification source.
For each extracted claim and its mapped source article, we apply Med-V1-L3B to predict a 5-point Likert score from strong contradiction ($-2$) and partial contradiction ($-1$) to neutral ($0$), partial agreement ($+1$), and strong agreement ($+2$). We define a claim as supported when the Med-V1 score is $+1$ or $+2$. Conditioned on successful PMID mapping, we define a content hallucination when the claim is not partially or fully supported by its mapped source, corresponding to Med-V1 scores of $-2$, $-1$, or $0$.

We report the following metrics by model and citation instruction: (i) the average number of extracted claims per answer, (ii) the PMID mapping rate used to identify citation fabrication, (iii) the average mapped PMID value as a proxy for recency, (iv) the content hallucination rate among successfully mapped citations, (v) the proportion of supported claims, and (vi) the number of supported claims. Human expert statistics are computed from the expert answers in MedAESQA using the same claim extraction pipeline. 

\subsection{Identifying High-Stakes Misattributions with Med-V1}
For this case study, we first compile a corpus of clinical guidelines by retrieving all the free full-text articles with the publication type of ``practice guideline'' and publication year between 2015 and 2025 (6,152 PMIDs, on Sep 16, 2025).
Then, we extract the citation statement sentences from these articles using the BioC-preprocessed version of PubMed Central \cite{comeau2019pmc}.
These articles have annotations indicating exactly where each citation appears in the text. 
Since some citation sentences might not be verification-worthy (e.g., citations to an experimental method, or a contextualized citation sentence that has ambiguous pronouns), we further apply GPT-4o-mini with Prompt~\ref{prompt:claim_quality} to filter out such claims, which leads to about 57k pairs of claim-citation pairs.
We apply Med-V1 on this corpus and take the subset of cases in which the model determines there is disagreement between the source and claim. Because the source is obtained from the guideline claim’s citations, any instance of disagreement is proposed to be a misattribution. To ensure a balanced sample across Med-V1’s two levels of disagreement (partial contradiction and strong contradiction), we first stratify the misattribution cases by output label, then draw a random sample of 50 cases from each group, totaling to 100 cases for analysis.

\begin{AIbox}{Claim Verification-worthy Check Prompt}
\label{prompt:claim_quality}
\textbf{System Prompt}\\
You are a biomedical expert, and your task is to classify if a biomedical claim can be fact-checked.\\
Please respond with ``yes'' if the claim meets the requirement, and ``no'' otherwise. Only output ``yes'' or ``no''.\\[0.2em]

\textbf{User Prompt}\\
Claim: ``\{claim\}''
\end{AIbox}

Our annotation procedure is similar to the one used for the error analysis. Two annotators independently review all 100 sampled cases. For each case, they evaluate, in order, the guideline claim, source (i.e. abstract of the article cited by the guideline), and Med-V1’s reasoning. As in the error analysis, they first classify cases with bad claims. Proceeding to the source, they recognize that a claim may be supported by evidence in the full text that is not present in the abstract. However, because Med-V1 uses titles and abstracts as input, we introduce a ``not enough information'' category to convey that this is due to lack of abstract detail, rather than a model error. When information in the abstract clearly does not support the claim, the case is labeled as a validated misattribution. On the other hand, if the abstract supports the claim, then Med-V1’s output is a model error. After completing their independent assessments, the annotators discuss disagreements and reach a final consensus category for each case. All reported validation counts and clinical-domain analyses use these consensus-adjudicated labels. We do not further subcategorize model errors, as this is addressed in the error analysis section. 

We perform a second-stage analysis on cases that are classified as validated misattributions, focusing on their potential clinical risk. We reuse the topic classification of clinical questions \cite{munn2018kind} to categorize claims into different domains/topics. Our classification results form the basis for further assessment of the potential risk associated with each misattribution. 

\section*{Data Availability}
MedFact-Synth, containing all 1,497,981 training instances, is publicly available at \url{https://huggingface.co/datasets/nlm-dir/MedFact-Synth}.
MedFact-Bench is available at \url{https://huggingface.co/datasets/nlm-dir/MedFact-Bench}. Source Data are provided with this paper.

PubMed 2025 baseline can be downloaded from \url{https://ftp.ncbi.nlm.nih.gov/pubmed/baseline/}.
Pre-processed SciFact and HealthVer can be downloaded from \url{https://github.com/dwadden/multivers}.
MedAESQA can be downloaded from \url{https://github.com/deepaknlp/MedAESQA}.
PubMedQA can be downloaded from \url{https://github.com/pubmedqa/pubmedqa}. 
BioASQ can be downloaded from \url{http://participants-area.bioasq.org/}.

\section*{Code Availability}
Our code is available at \url{https://github.com/NLM-DIR/Med-V1}. The Med-V1-L3B model is available at \url{https://huggingface.co/nlm-dir/Med-V1-L3B}, and the Med-V1-Q3B model is available at \url{https://huggingface.co/nlm-dir/Med-V1-Q3B}.

Llama-3.2-3B-Instruct is available at \url{https://huggingface.co/meta-llama/Llama-3.2-3B-Instruct}. Qwen2.5-3B-Instruct is available at \url{https://huggingface.co/Qwen/Qwen2.5-3B-Instruct}. We access GPT-5, GPT-4, GPT-4o, o3-mini, and Llama-3.3-70B-Instruct through the application programming interface (API) provided by Microsoft Azure \url{https://azure.microsoft.com/}, and generation temperature is set to 0 wherever applicable for deterministic outputs.

\section*{Author Contributions}
Q.J. conceived the study. Q.J. and Z.L. designed the experiments. Q.J., Y.F., Y.Y., and D.C. conducted data collection, model training, and evaluation. Z.W., N.W., J.C., R.L., and C.F. performed data annotations and error analysis. L.H., N.W., C.F., G.X., A.Z., M.C., and Y.P. analyzed case studies. Z.L. supervised the project. All authors contributed to writing the manuscript and approved the submitted version.

\section*{Acknowledgements}
This research was supported by the Intramural Research Program of the National Institutes of Health (NIH). The contributions of the NIH author(s) are considered Works of the United States Government. This research was also partially supported by the NIH Pathway to Independence Award K99LM014903 (Q.J.), as well as R01LM014344 (Y.P.) and R01LM014573 (Y.P.). The findings and conclusions presented in this paper are those of the author(s) and do not necessarily reflect the views of the NIH or the U.S. Department of Health and Human Services.

\section*{Competing interests}
None declared.

\newpage

\bibliography{sn-bibliography}

\end{document}